\documentclass[letterpaper]{article} 
\usepackage{aaai24}  
\usepackage{times}  
\usepackage{helvet}  
\usepackage{courier}  
\usepackage[hyphens]{url}  
\usepackage{graphicx} 
\usepackage{natbib}  
\usepackage{caption} 
\usepackage{algorithm}
\usepackage{algorithmic}
\usepackage{mathtools}
\usepackage{multirow}
\usepackage{booktabs}
\usepackage{bm}
\usepackage{newfloat}
\usepackage{listings}
\DeclareCaptionStyle{ruled}{labelfont=normalfont,labelsep=colon,strut=off} 
\floatstyle{ruled}
\newfloat{listing}{tb}{lst}{}
\floatname{listing}{Listing}
\usepackage{color}
\usepackage{color}
\definecolor{shadecolor}{rgb}{0.92,0.92,0.92}  
\usepackage{framed} 

\title{Self-driven Grounding: Large Language Model Agents with Automatical Language-aligned Skill Learning}
\author{
    Shaohui Peng\textsuperscript{\rm 1},
    Xing Hu\textsuperscript{\rm 2},
    Qi Yi\textsuperscript{\rm 2, \rm 5},
    Rui Zhang\textsuperscript{\rm 2, \rm 4},
    Jiaming Guo\textsuperscript{\rm 2},
    Di Huang\textsuperscript{\rm 2}, \\
    Zikang Tian\textsuperscript{\rm 1},
    Ruizhi Chen\textsuperscript{\rm 1, \rm 3},
    Zidong Du\textsuperscript{\rm 2, \rm 4},
    Qi Guo\textsuperscript{\rm 2},
    Yunji Chen\textsuperscript{\rm 2, \rm 3},
    Ling Li\textsuperscript{\rm 1, \rm 3}
}
\affiliations{
    \textsuperscript{\rm 1} Intelligent Software Research Center, Institute of Software, CAS\\
    \textsuperscript{\rm 2} SKL of Processors, Institute of Computing Technology, CAS\\
    \textsuperscript{\rm 3} University of Chinese Academy of Sciences \\
    \textsuperscript{\rm 4} Cambricon Technologies\\
    \textsuperscript{\rm 5} University of Science and Technology of China \\
    
    {pengshaohui@iscas.ac.cn}
}

\usepackage{bibentry}

\begin{document}

\maketitle

\begin{abstract}

Large language models (LLMs) show their powerful automatic reasoning and planning capability with a wealth of semantic knowledge about the human world. However, the grounding problem still hinders the applications of LLMs in the real-world environment. Existing studies try to fine-tune the LLM or utilize pre-defined behavior APIs to bridge the LLMs and the environment, which not only costs huge human efforts to customize for every single task but also weakens the generality strengths of LLMs. To autonomously ground the LLM onto the environment, we proposed the Self-Driven Grounding (SDG) framework to automatically and progressively ground the LLM with self-driven skill learning. SDG first employs the LLM to propose the hypothesis of sub-goals to achieve tasks and then verify the feasibility of the hypothesis via interacting with the underlying environment. Once verified, SDG can then learn generalized skills with the guidance of these successfully grounded subgoals. These skills can be further utilized to accomplish more complex tasks which fail to pass the verification phase. Verified in the famous instruction following task set-BabyAI, SDG achieves comparable performance in the most challenging tasks compared with imitation learning methods that cost millions of demonstrations, proving the effectiveness of learned skills and showing the feasibility and efficiency of our framework.
\end{abstract}
\newcommand{\instf}{instruction following}
\newcommand{\alg}{ALG\_NAME}
\newcommand{\yq}[1]{{\color{red} (YQ: #1)}}
\newcommand{\xing}[1]{{\color{blue} #1}}
\newcommand{\xingNote}[1]{{\color{red} (X: #1)}}
\newcommand{\psh}[1]{{\color{blue} (PSH: #1)}}
\newcommand{\di}[1]{{\color{blue} di: #1}}
\newcommand{\xname}{}

\section{1 Introduction}

Large language models (LLMs) have shown their powerful capability in automatic reasoning and planning with a wealth of semantic knowledge about the human world \cite{LLM1,LLM2,LLM3,LLM4}. 
However, there still remains a large gap in adopting LLMs to automatically solve problems in specific environments. This is because of the misalignment between the LLM's semantic planning and the grounded-specific implementation, which is also known as the grounding problem \cite{saycan,ground1}. 
Solving this problem can unlock the LLMs' capacity of understanding and affecting the real world, which is a solid step towards real-world applications of artificial intelligence.




%

To address the grounding problem, existing studies try to fine-tune the LLM to predict feasible actions \cite{llm_finetune, llm_ft1, llm_ft2} or utilize a set of behavior APIs (i.e. low-level skills) that serve as a bridge between the LLMs and the environment \cite{saycan, reprompt, codepol}.
On the one hand, fine-tuning LLMs is of low sample efficiency and may also damage the reasoning ability of LLMs. 
On the other hand, existing methods relying on behavior APIs often assume the APIs are pre-defined by the environment \cite{codepol} or pre-trained using expert demonstrations \cite{saycan}, which not only costs huge human efforts to customize for every single task but also weakens the generality strengths of LLMs. 
Therefore, how to \emph{autonomously } ground the LLM onto the environment still remains an open problem and is the key challenge of LLM-based agents.

It is challenging to achieve the goal of autonomous grounding that maps the LLM's semantic plan to practical implementation, because of the following reasons: 1) The prerequisite of grounding, obtaining successful experiences, is difficult because of the sparse rewards in the physical world. 2) Even obtaining rare success experiences, the grounding is usually closely related to specific tasks without a shared API library, therefore is of low generality and invaluable for general tasks. 
To address these issues, we produce intrinsic rewards based on LLM-generated subgoals and their check functions, which increase successful experiences by alleviating the sparse reward issue. We then propose the language-aligned general skill learning methodology by forcing each skill to achieve a group of goals with similar semantic descriptions. These skills show good generality in solving other or even more complex tasks. 

In summary, we propose a Self-Driven Grounding (SDG) framework that intimately combines the LLM and the reinforcement learning process within the following key stages: 1) Hypothesis: the LLM not only acts as the planner by decomposing tasks into small subgoals but also provides the check functions so that RL agents can evaluate whether they can complete these subgoals. Such intrinsic rewards from the LLM significantly alleviate the sparse reward issue. 2) Verification: with the subgoals and corresponding check functions, RL agents learn the policies of the subgoals based on intrinsic rewards and finally are verified through whether the tasks are accomplished.  3) Induction: RL agents cluster verified subgoals through semantic similarity and learn generalized skill policy upon them. With these general skills, the LLM can generate solutions for unseen or even more complex tasks through minimal and efficient interaction.  

We validate the self-driven grounding framework in instruction following tasks, which is common and reasonable for LLM-based agents because of textural task instruction. Verified in BabyAI, a grid world platform to study language-grounded tasks, our automatic grounding framework achieves comparable performance in the most difficult tasks compared with imitation learning methods that cost millions of demonstrations.
The experiment results not only proves the effectiveness of learned skills but also show the feasibility and efficiency of our framework. 

\section{2 Related Work}

\paragraph{LLM-assisted Agents}
LLMs show their great power in automatic reasoning and planning with a wealth of semantic knowledge about the human world. 
Therefore, it is promising to involve LLMs in developing intelligent agents.
The key challenge in LLM-assisted agents is how to ground the LLM’s knowledge (in linguistic form)
to the tasks at hand.
Regarding this challenge, there are two mainstream methods:
(1) utilizing a set of behaviour APIs with detailed linguistic annotations. 
Such APIs can be pre-defined by the environment \cite{codepol} or pre-trained using expert demonstrations \cite{saycan,innermono,plan4mc}. 
For example, Code as Policies \cite{codepol} uses the LLM to generate executable codes for accomplishing instructions which can invoke behaviour APIs under certain conditions.
SayCan \cite{saycan} invites humans to rate the success of given demonstrations, which are utilized to train API policies and then derive an affordance function.
Voyager \cite{voyager} stores and retrieves executable code, which calls pre-implemented basic APIs, to handle complex scenarios. 
To better interact with the environment, some approaches \cite{innermono,reprompt} introduce the environment feedback to re-generate new plans, which can be seen as another kind of trial-and-error learning. 
Although these methods have made impressive progress by utilizing APIs, their applications are also limited by the behaviour APIs in that the agent can only accomplish tasks that can be solved by arranging these basic APIs. 
(2) fine-tuning the LLMs. The LLM can be fine-tuned to predict the agent's feasible action given the state descriptions. Such fine-tuning can be performed using expert demonstrations \cite{llm_ft2, llm_ft1} or online RL \cite{llm_finetune}. However, fine-tuning a model as large as the LLM is quite time expensive and requires much training data.


\paragraph{\textbf{Instruction Following}}
In \instf{}, an agent is given an instruction and the goal is to accomplish the task described by the instruction.
Such a paradigm makes the agent able to assist human beings by following human instructions, which has wide real-world applications.
The works for \instf{} can be divided into three categories:
(1) Semantic-parsing methods \cite{iw0,iw1}, which directly parses the instruction into the agent's actions via lexical analysis and other pre-defined rules. These methods require great human efforts to design proper rules, and can not generalize to complex environments.
(2) Learning-based methods, which directly train a language-conditioned policy to accomplish instructions. 
Many prior works require expert demonstrations in their training loops.
For example, expert demonstrations are often used in policy imitation learning \cite{rwil1, rwil2}, hindsight instruction relabelling \cite{rwhr1,rwhr2}, and learning the language-conditioned reward function \cite{rwirl1}.
Some works try to sidestep the need for expert demonstrations \cite{ella, higher}, but at the cost of much lower sample efficiency.
All learning-based approaches are typically trained using hard-coded instruction templates, which can not provide diverse, ambiguous and long-term planning instructions as humans. 
Therefore, they can only deal with simple and low-level instructions such as pick-and-place tasks.
(3) LLM-based methods, which use LLMs to assist the understanding and planning of instructions \cite{saycan, codepol, reprompt}. 
See the last paragraph for more details.

\section{3 Preliminaries}
\subsection{3.1 Problem Formulation}
We consider adopting an LLM-based agent to solve instruction following (IF) tasks.
Each instruction $I \in T$ describes a task coarsely in the environment.
Given the instruction, only when the agent accomplishes the task using primitive action set $A$ can receive a positive reward from the environment.
For example, in BabyAI, which is a famous instruction following task set, instructions like ``Open the green door'' or ``Put the red box next to the blue ball'' specify some macroscopic object manipulation tasks, while the agent needs to accomplish them in a grid world using primitive actions like ``turn right'', ``move forward'', ``pick'' and so on.

\begin{figure}[tb]
\centering
\includegraphics[width=0.45\textwidth]{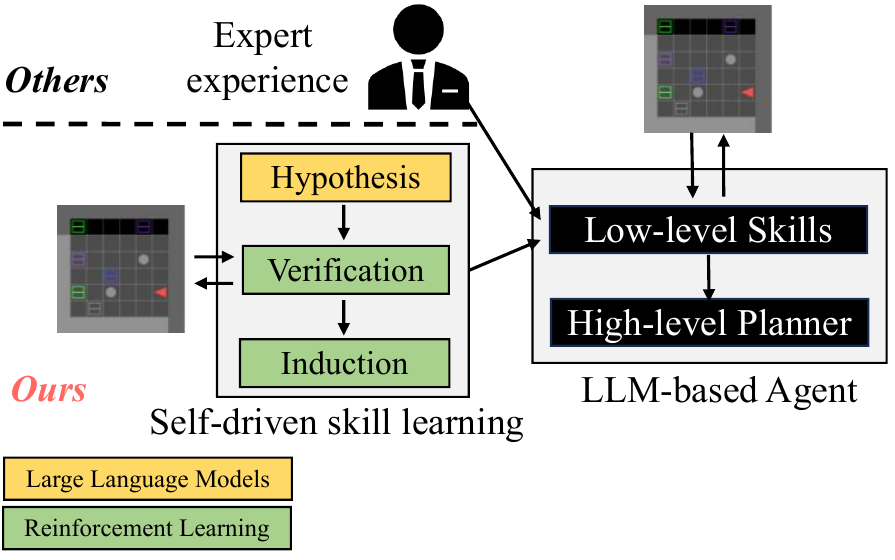}
\caption{A general framework of LLM-based agents}
\label{fig_comparison}
\end{figure}

An LLM-based agent takes instructions $I$ and environment observation $o$ as input and outputs actions to accomplish tasks.
As shown in Figure \ref{fig_comparison}, the general framework of LLM-based agents contains a high-level \textit{planner} and some low-level \textit{skills} that can unitize the semantic knowledge in the LLM to accomplish instruction following tasks.
Given the coarse instruction, the \textit{planner} (often LLMs) will decompose it into a sub-instruction sequence or generate a program to solve it.
At the same time, the low-level skills consist of pre-trained policies or pre-implemented scripts to execute the plan or program.
Researchers usually assume the environment provides textual descriptions of state and task-related feedback to adapt to the LLM setting.
With the immediate translation from the semantic output to execution in the environment through low-level APIs, the high-level planner can get adequate feedback and iteratively refine its plan to accomplish tasks.

Current methods leverage semantic knowledge in the high-level planner to reason and decompose the coarse instruction, but meanwhile, bypass the grounding problem by translating the semantic plan into implementation through pre-defined low-level APIs.
In order to maximize the use of priors in LLMs to reduce human effort, our framework aims to \textit{automatically learn generalized skills in the environment to build low-level APIs to solve the grounding problem}.
Besides, in this paper, we assume the LLM could call basic perception functions $P=\{p1,p2,...\}$, like ``get\_observed\_objects()'', to get environment status instead of pre-designed textual observation and feedback mechanisms.

\subsection{3.2 Challenges}
The essential problem of building an LLM-based agent is to ground semantic knowledge of LLM in the environment. 
To automatically solve the grounding problem,
there are two main challenges:
\begin{itemize}
    \item How to obtain successful grounding experiences from scratch? 
    Without pre-defined low-level APIs, the agent cannot interact with the environment to attempt the semantic plan directly and efficiently explore the reward of task accomplishment.
    To address the challenge, we make the LLM hypothesize the plan for each task and generate corresponding checks for each sub-step.
    Based on these intrinsic rewards provided by checks, we can quickly train small policies to execute and verify the plan, then collect successful trajectories as grounding experiences. 
    Although such experience obtained through quick attempts may belong to simple tasks, we can also leverage them to enhance the grounding ability of the agent to accomplish more complex and long-term tasks progressively.
    \item How to efficiently train generalized low-level behavior APIs under the guidance of experience?
    The subgoals in successful experience are proposed by the LLM based on specific instruction, and cannot be applied to new scenarios as general behavior APIs.
    Inspired by inductive ideas in mathematics, we introduce a mechanism to group subgoals with similar semantics together, then train skill policies that can achieve a group of subgoals as generalized behavior APIs.
    Unlike standard online reinforcement learning, which suffers from data efficiency issues, we make skills training efficient through dense rewards provided by checks, and initial state restoration by the successful trajectories.
\end{itemize}





\section{4 Method}
In this section, we will first give an overview of our proposed Self-Driven Grounding framework, which can automatically and progressively ground LLM in the environment. 

\subsection{4.1 Overview}
\begin{figure*}[tb]
\centering
\includegraphics[width=0.95\textwidth]{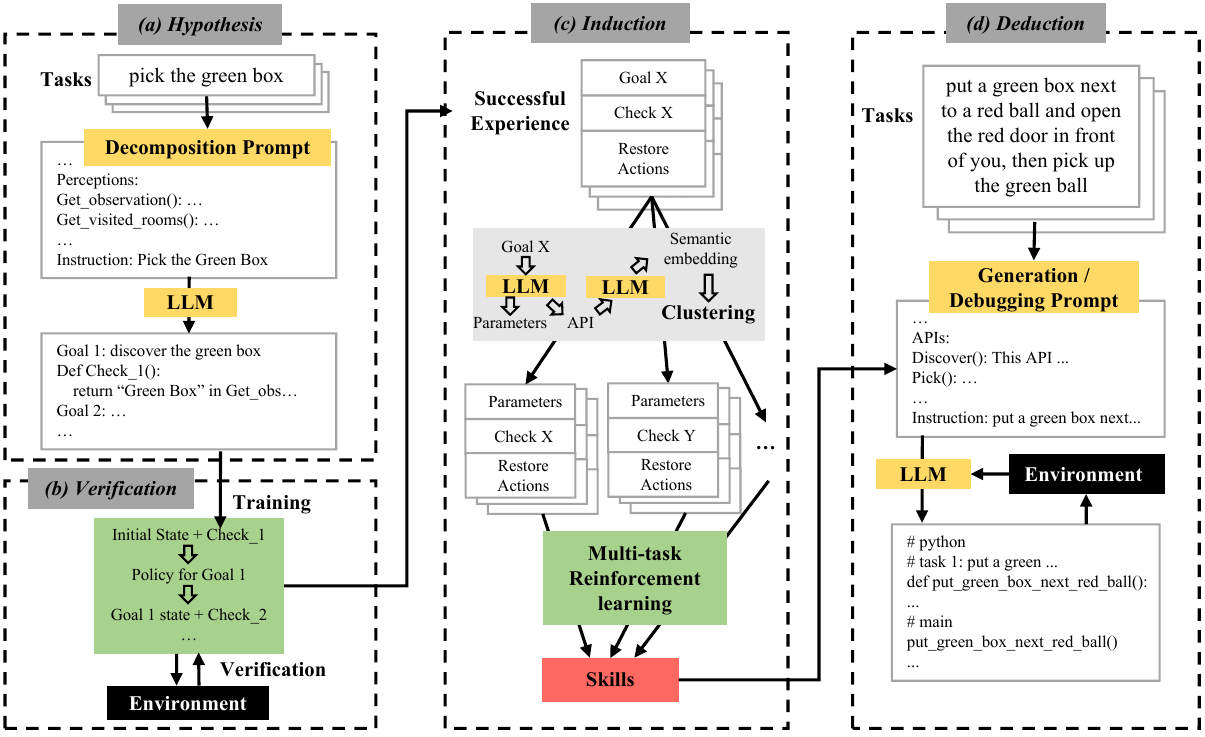}
\caption{Overview of Self-Driven Grounding (SDG) framework. (a) Hypothesis: try to decompose tasks; (b) Verification: leverage efficient interaction with the environment to verify the results of hypothesis; (c) Induction: group successful experience to train generalized skills; (d) Deduction: build few-shot planner to solve tasks using acquired skills. }
\label{fig_overview}
\end{figure*}

As shown in Figure \ref{fig_overview}, SDG can be divided into four phases.
\begin{itemize}
    \item \textbf{Hypothesis}: For each instruction, LLM tries to decompose it into subgoals and generate check functions for each subgoal. 
    \item \textbf{Verification}: Based on the reward provided by the check function, we train separate policies for each subgoal within limited steps until the task accomplishes to verify the feasibility of the hypothesis of LLM.
    \item \textbf{Induction}: We group the subgoals in successful hypotheses with similar semantics to train generalized skills reinforcement learning.
    \item \textbf{Deduction}: Based on learned skills as low-level actors, we use LLM as a few-shot high-level planner to generate programs to solve unseen and more complex tasks.
\end{itemize}

\subsection{4.2 Hypothesis}
The hypothesis phase aims to solve tasks separately regardless of generality to collect grounding experience.
Considering the gap between the semantic knowledge in LLM and the environment, the hypothesis phase decomposes the task into several subgoals rather than directly giving the solution, and leaves correctness verification to the next process.
As shown in Figure\ref{fig_overview}(a), the hypothesis can be formed as $I \xrightarrow{Prompt}G,F$.
We use LLM as the zero-shot planner, which takes an instruction $I$ and necessary decomposition prompt as input, then outputs a subgoal sequence $G = \{g_1, g_2,...\} $ and corresponding check functions $F = \{f_1, f_2, ...\}$.
The subgoal $g_i$ is a small instruction labeled with the explicit mark ``Goal X'' to facilitate further processing.
Each check function $f_i: S\rightarrow \{0,1\}$ is a program that checks the achievement of corresponding sub-goal $g_i$ via invoking perception functions provided by the environment.
To make the LLM output as we want, except for the task instruction $I$, we add role definition, perception API descriptions, and explanation of the instruction following task space to the decomposition prompts.

\subsection{4.3 Verification} 
After getting the subgoals and check functions, we need to verify their feasibility in the environment to collect successful grounding experiences.
The feasibility of decomposition is verified by the consistency between achieved signals generated by check functions and the tasks accomplished signal.
Specifically, as shown in Figure \ref{fig_overview}(b), we train independent policies for each subgoal based on the bool reward provided by its check function.
Once the subgoal is achieved, we stop the training and save the action sequence.
The saved action sequence can be used as the restoring mechanism to prepare the initial state for further skill training.
Until all sub-goals are achieved and the task is also accomplished, we can verify the decomposition is successful.
Meanwhile, the grounding experiences (including subgoal descriptions, check functions, and restore action sequences) are collected for skill learning in the future phase.
Considering some complex and long-term tasks cannot be solved by direct decomposition, the above mechanism is only suitable for solving simple tasks within an acceptable environment interaction steps to collect experience.
We set the maximum number of verification steps $T_{verify}$ as a threshold to distinguish intractable complex tasks for the next stage to solve.

\subsection{4.4 Induction}
After collecting successful grounding experiences, the induction phase aims to discover and learn generalized skills from separate grounding trajectories of different instructions so that we can reuse them in more unseen and complex tasks.
\paragraph{Discovery}
As described above, we have collected task-solving experience through efficient hypothesis and verification, including subgoal descriptions, corresponding check functions, and start state restore action sequences. 
However, such successful experiences can only be used on specific instructions that are easy to decompose for LLM. 
To make the LLM-based agent able to solve more unseen and complex tasks, we must further abstract and learn skills to build the generalized low-level actor.
To this end, we cluster the collected subgoals according to their semantics to ensure a certain generalized skill can accomplish a category of subgoals as shown in Figure \ref{fig_overview}(c). 
Specifically, we first use LLM to translate each subgoal description $g_i$ into API description $g_{i,api}$ and parameter $g_{i, param}$.
For example, the subgoal ``discover the green box'' are translated into the API ``discover'' and parameter ``green box''.
Then we use the k-means algorithm to conduct unsupervised clustering based on the semantic distance between subgoal descriptions computed by the following cosine similarity:
\begin{equation}
    C(g_{i,api},g_{j,api}) = \frac{emb(g_{i,api}) \cdot emb(g_{j,api})}{||emb(g_{i,api})||\cdot ||emb(g_{j,api})||},
\end{equation}
where $g_{i, api}$ is the API description and $emb(\cdot)$ is the embedding functions of LLM.
\paragraph{Training}
We have divided subgoals with similar semantics into different categories through the clustering process.
Then we build reinforcement learning (RL) environments to train skills that can achieve a cluster of subgoals separately.
Unlike common RL environments, the skill training environment is like a multitask learning scenario.
Each subgoal of a cluster can be seen as a single task, the subgoal parameter $g_{i,param}$ is the task description, the reward is provided by the corresponding check function $f_i$, and the initial state is set by the saved restore action sequence.
Trained in such a multiple tasks environment consists of subgoals that have the same semantics, the skill is supposed to have generalization in tasks with similar scenarios. 
Besides, subgoals belonging to the same cluster are divided into training and verification sets to monitor the generalization ability of the trained skill to prevent overfitting.

\subsection{4.5 Deduction}
Through the above process, we have overcome the challenges of obtaining successful grounding experience from scratch and training generalized skills efficiently.
In other words, we have autonomously built the low-level skills for the LLM-based agent without human effort.
To apply automatically learned skills to solve unseen and complex instructions following tasks, we next introduce the high-level planner in this section.
As shown in Figure \ref{fig_overview}(d), we use LLM as a few-shot planner to generate programs to accomplish tasks.
The process can be divided into program generation phase and debugging phase.
\paragraph{Generation} 
The generation prompt for LLM contains the role definition and API descriptions (including skills and perception functions).
Besides, considering the complexity of the task, the generation prompt follows a few-shot in-context learning paradigm.
We include some skill API descriptions and a hand-written example that leverages the learned skills to solve a complex instruction-following task.
\paragraph{Debugging} To better solve complex tasks, we also designed an interaction debugging process in our high-level planner.
Besides the task instruction and the generated program to be modified, the debugging prompt also includes the error message and some general debugging suggestions to fix the possible bugs.
Benefiting from the feasibility and robustness of adaptive learned skills, the basic error reporting mechanism based on illegal action detection can effectively improve the accuracy of the generated programs, which greatly reduces human effort.


\begin{figure}[tb]
\centering
\includegraphics[width=0.48\textwidth]{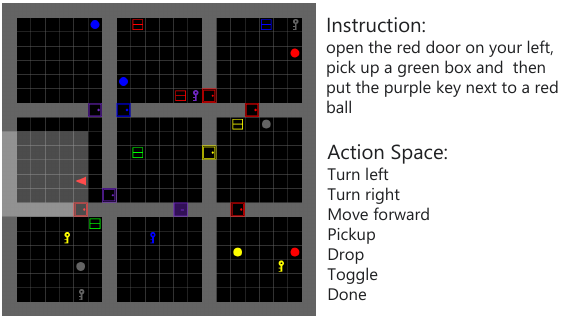}
\caption{A example of BossLevel task in BabyAI}
\label{fig_baby}
\end{figure}
\begin{table}[tb]
\centering
\begin{tabular}{ccccc}
\hline
Task                          & Method     & Type & Demos & rate \\ \hline
\multirow{4}{*}{GoToLocal}    & Original   & IL   & 10K   & 99.8 \\
                              & LID-Text   & IL   & 10K   & 99.5 \\
                              & SDG-action & LLM  & 0     & 55.1 \\
                              & SDG        & LLM  & 0     & 99.9 \\ \hline
\multirow{4}{*}{PickupLoc}    & Original   & IL   & 10K   & 99.8 \\
                              & LID-Text   & IL   & 10K   & 99.8 \\
                              & SDG-action & LLM  & 0     & 47.6 \\
                              & SDG        & LLM  & 0     & 92.9 \\ \hline
\multirow{4}{*}{PutNextLocal} & Original   & IL   & 10K   & 97.7 \\
                              & LID-Text   & IL   & 10K   & 99.9 \\
                              & SDG-action & LLM  & 0     & 0    \\
                              & SDG        & LLM  & 0     & 91.9 \\ \hline
\multirow{3}{*}{Open}         & Original   & IL   & 1M    & 100  \\
                              & SDG-action & LLM  & 0     & 0    \\
                              & SDG        & LLM  & 0     & 92.4 \\ \hline
\multirow{4}{*}{SynthSeq}     & Original   & IL   & 1M    & 87.7 \\
                              & LISA       & IL   & 100K  & 61.2 \\
                              & SDG-action & LLM  & 0     & 0    \\
                              & SDG        & LLM  & 0     & 78.9 \\ \hline
\multirow{4}{*}{BossLevel}    & Original   & IL   & 1M    & 84.3 \\
                              & LISA       & IL   & 100K  & 69.8 \\
                              & SDG-action & LLM  & 0     & 0    \\
                              & SDG        & LLM  & 0     & 75.9 \\ \hline
\end{tabular}
\caption{Overall results. ``IL'' means ``Imitation Learning'', ``LLM'' means ``LLM-based agent''.}
\label{tb:overall}
\end{table}
\begin{figure}[tb]
\centering
\includegraphics[width=0.48\textwidth]{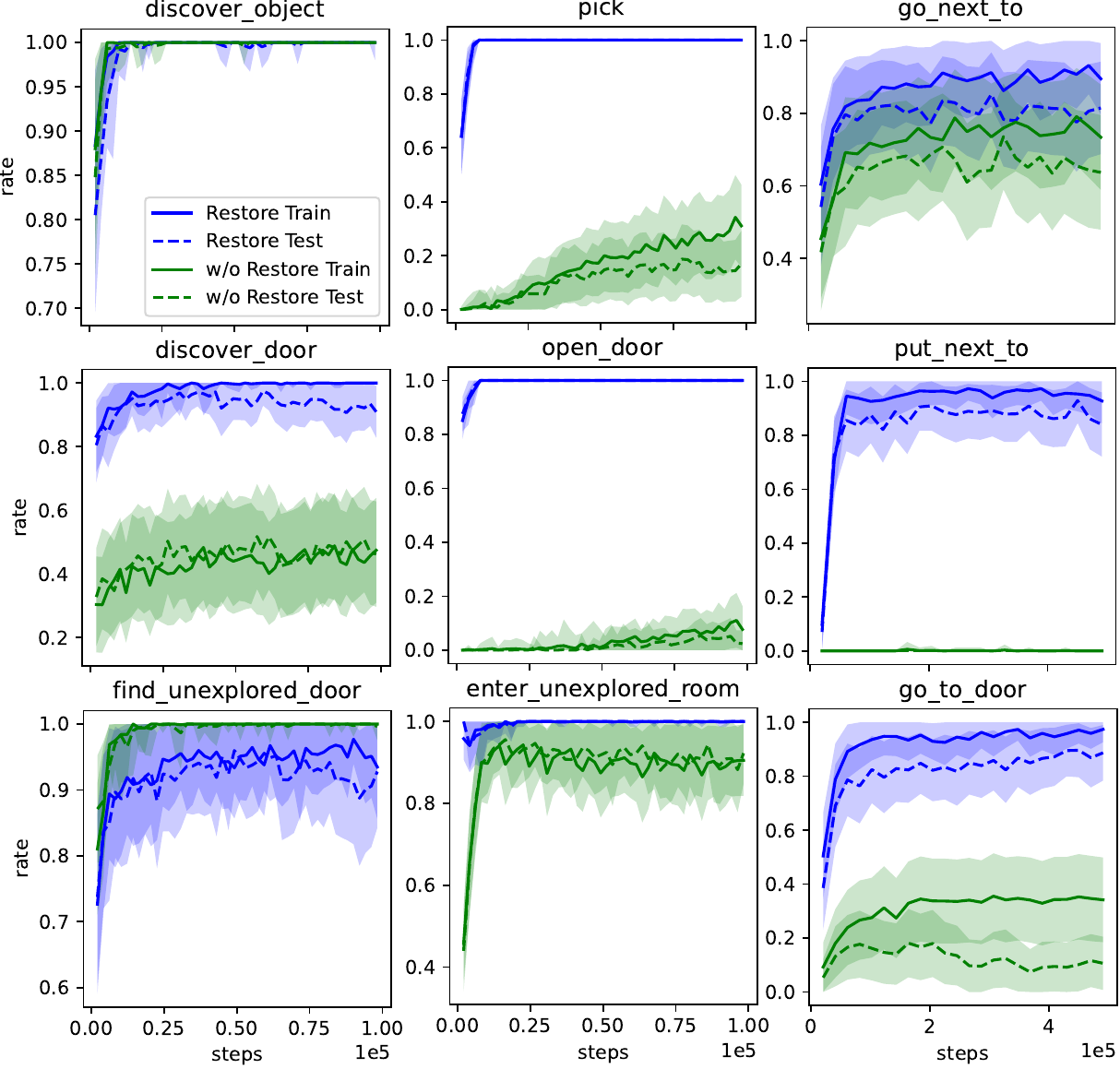}
\caption{Training curves of skills. The blue curve represents skill learning through restoring the initial state, while the green curve didn't. The dotted curve represents test performance in the held-out verification set.}
\label{fig_skill}
\end{figure}
\begin{figure}[tb]
\centering
\includegraphics[width=0.45\textwidth]{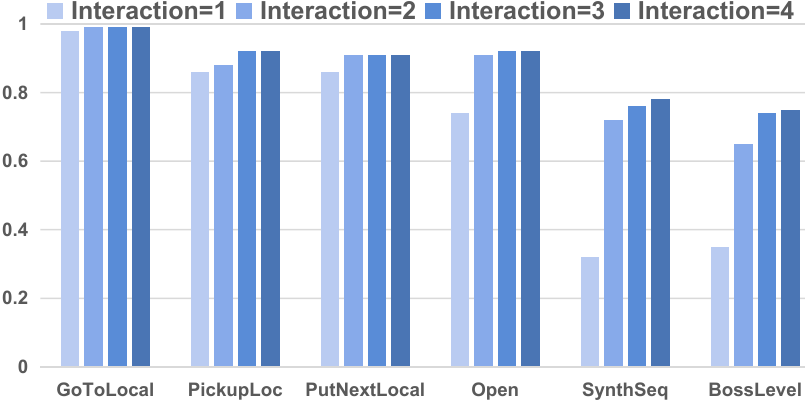}
\caption{Successful rates of different interaction times}
\label{fig_inter}
\end{figure}
\begin{figure}[tb]
\centering
\includegraphics[width=0.45\textwidth]{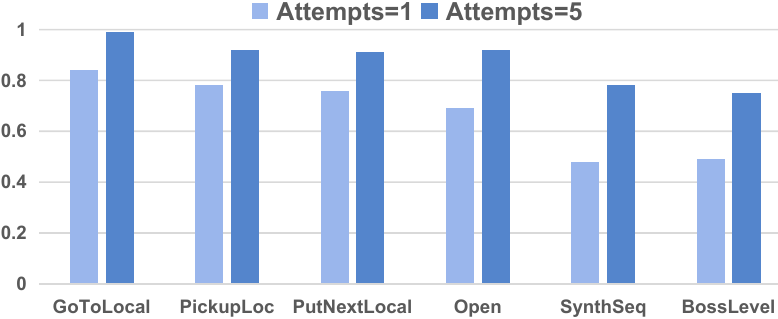}
\caption{Successful rates of different attempts times}
\label{fig_attempts}
\end{figure}
\begin{figure}[tb]
\centering
\includegraphics[width=0.3\textwidth]{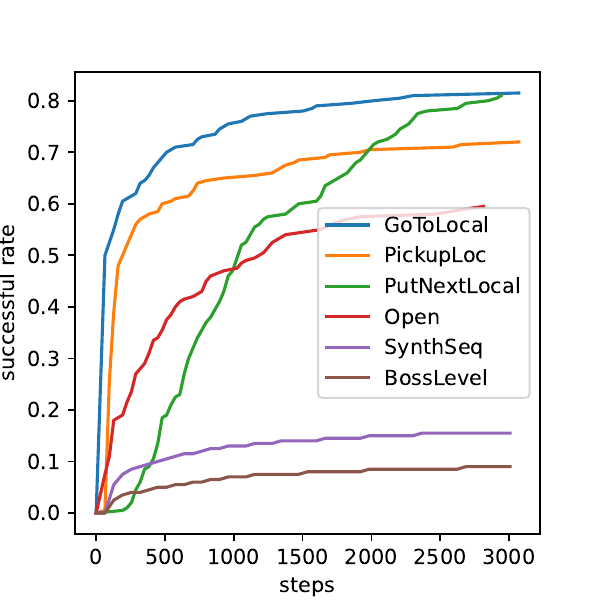}
\caption{Task verification results}
\label{fig_verify}
\end{figure}
\section{5 Results}
\subsection{5.1 Experiment Setting}
\subsubsection{Environment}
To evaluate the efficiency and effectiveness of our proposed framework that automatically discovers, learns, and applies skills, we test SDG on the BabyAI environment \cite{babyai}.
BabyAI is a grid world environment for instruction following.
Given the language instruction and a $7\times 7\times 3$ partial and local view, the agent must learn to accomplish various tasks of arbitrary difficulty levels.
In this paper, we choose the following six levels of instruction with different types and difficulties (more details of the environment can be found in the Appendix):
\begin{itemize}
    \item \textbf{GoToLocal}: Go to an object inside a single room.
    \item \textbf{PickupLocal}: Pick up an object inside a single room.
    \item \textbf{PutNextLocal}: Pick up an object and put it next to another object inside a single room.
    \item \textbf{Open}: Open a door in a $3\times 3$ room maze, the door may in another room.
    \item \textbf{SynthSeq}: Union of all instructions from PutNext, Open, Goto, and PickUp and may with multiple commands.
    \item \textbf{BossLevel}: The hardest task of BabyAI as shown in \ref{fig_baby}. The command can be any sentence drawn from the Baby Language grammar.
\end{itemize}
\subsubsection{Baselines}
We verify the effectiveness of SDG by comparing it with several baselines, including Imitation Learning methods relying on expert demonstrations and a variant of our LLM-based agent. 
\begin{itemize}
    \item \textbf{Original}:  The baseline from the original BabyAI paper, which trained the GRU + CONV model with imitation learning using one million demonstration episodes for each level.
    \item \textbf{LID-Text \cite{llm_ft1}}: An approach that first represents goals and observations as a sequence of embeddings, then uses a policy network initialized with a pre-trained LM and trained with demonstrations to predict the next action.
    \item \textbf{LISA \cite{LISA}}: A hierarchical imitation learning framework that can learn diverse, interpretable skills from language-conditioned demonstrations.
    \item \textbf{SDG-action}: The variant of our framework which maintains the same high-level planner but employs primitive actions instead of the acquired skills.
\end{itemize}
\paragraph{Implementation}
In this paper, we use ChatGPT (GPT-3.5-turbo) as the large language model to complete task decomposition, the semantic embedding of API, high-level planning, and debugging.
More details on the prompt contents are shown in the Appendix.
In the verification and skill learning phase, we use the standard model proposed in BabyAI, and train the policy using the PPO algorithm.

\subsection{5.2 Overall Results Comparison}
The main performance results are shown in Table \ref{tb:overall}.
We separately compare SDG with baselines in each level task.
For each level task, we randomly sample 100 instructions that never occurred in the skill training phase.
Considering the randomness of ChatGPT's answers, we repeat the experiment of each instruction 3 times to get the average results.
The results showed that SDG can achieve comparable performance using automatically learned skills rather than a large number of expert demonstrations, which shows the effectiveness of our framework.
Besides, the results of SDG-action showed it failed to accomplish harder tasks without acquired skill, which also emphasizes the importance of self-driven skill learning in the LLM grounding scenario.

\subsection{5.3 Ablation Results}
\subsubsection{Skill Learning Ablation}
We first investigate the data efficiency when learning skills through reinforcement learning as shown in Figure \ref{fig_skill} .
Although the difficulty varies because the clustering process only relies on the semantics of subgoals and ignored the difficulty, skills can be learned efficiently.
The verification phase proves that the instruction has been decomposed into small-enough subgoals and can be trained in limited steps.
Therefore, the skill training environment consisting of verified subgoals can lead to an efficient skill training process.
Besides, results also demonstrate that restoring actions that make the skill start at an expected state is important to learning efficiency.
Without the start state reset, the learning efficiency drops obviously, and some skills even cannot be learned.
Some learning curves show overfitting trends in the late period, which is different from the normal reinforcement learning process and Illustrates the role of the hold-out validation set.
For some skills like ``enter\_unexplored\_room'', the green curves seem better than the blue ones.
This is because without restoring the initial state, the difficulty decreases a lot since the agent can enter any room.


\subsubsection{Deduction Ablation}
We also explored the effect of the interaction debugging times and multiple attempts of skills. 
The ablation results show the importance of interactive debugging and multiple attempts.
\paragraph{A. Interaction Times:} Figure \ref{fig_inter} shows the results of different interaction times between the high-level planner and the environment.
For some tasks like ``GoToLocal'', the successful rate promotion is limited because of simplicity.
For complex tasks like ``BossLevel'', repeat debugging can bring more than $40\%$ promotion, which shows the adaptability and feasibility of learned skills.
However, when the performance reaches some ceiling bound, more interaction seems useless.

\paragraph{B. Multiple Attempts:} Different from pre-defined APIs with scripts, our learned skills are stochastic policies.
Thus, we also investigate the effect of multiple attempts of skill policy on the final success rate.
Figure \ref{fig_attempts} shows similar results with interaction times, the multiple attempts improve complex tasks greater.

\subsection{5.4 Method Details}
To show more insight into SDG, we show some key intermediate results.
\paragraph{Task Verification}
Figure \ref{fig_verify} shows the verification results of different level tasks.
In the implementation, we random sample 100 instructions from each level task, and set the verification steps threshold $T_{verify}$ equals to 3000.
The results prove our assumption, for some simple task levels, like ``GoToLocal'', ``PickupLoc'' and ``PutNextLocal'', the LLM can decompose the instruction into reasonable subgoals and check functions, so that the verification training be successful in limited steps.
For hard levels which also include some simple tasks, direct decomposition can accomplish few instructions, which means it cannot solve complex and long-term tasks.

\begin{figure}[tb]
\centering
\includegraphics[width=0.35\textwidth]{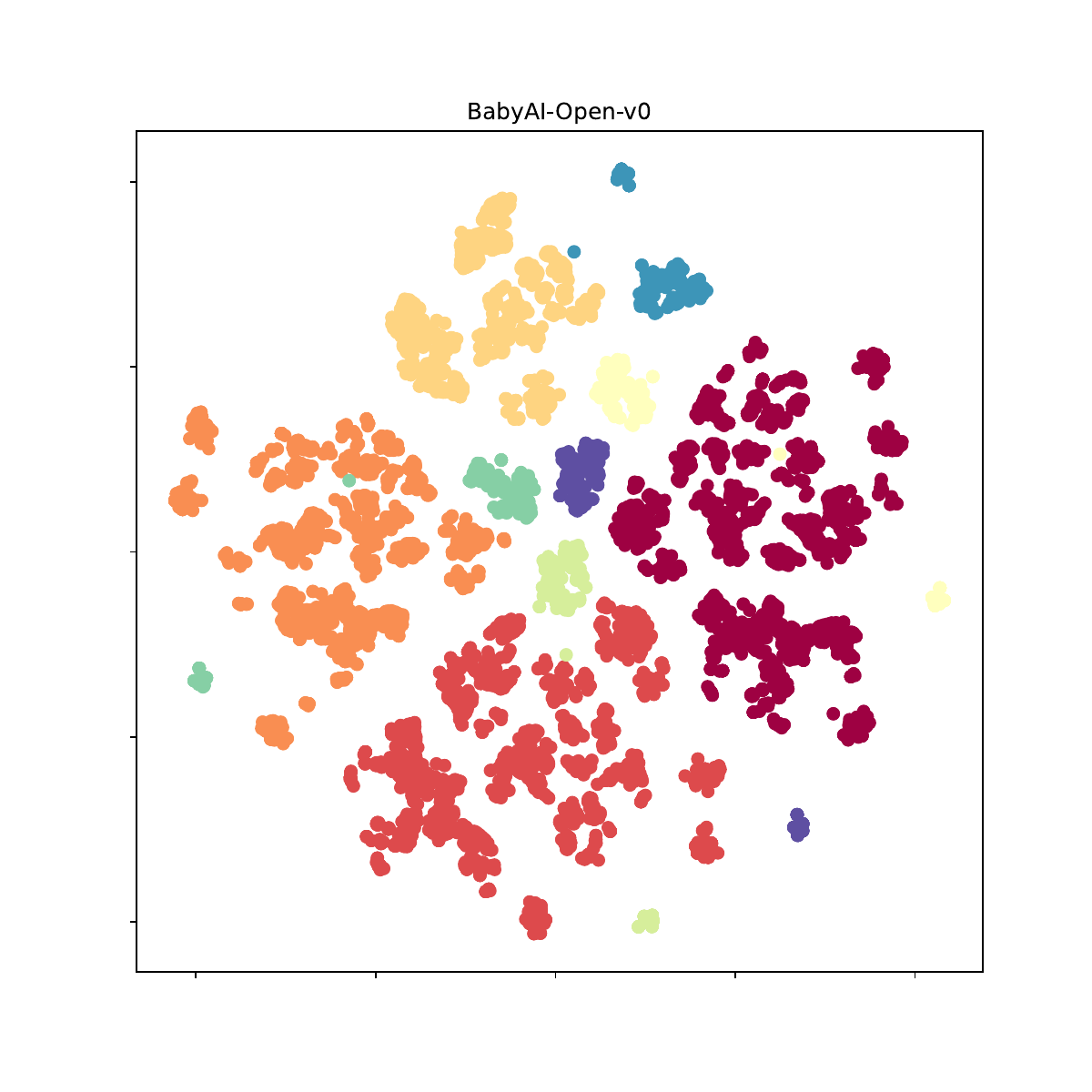}
\caption{Task clustering results}
\label{fig_ch}
\end{figure}

\begin{figure}[tb]
\centering
\includegraphics[width=0.43\textwidth]{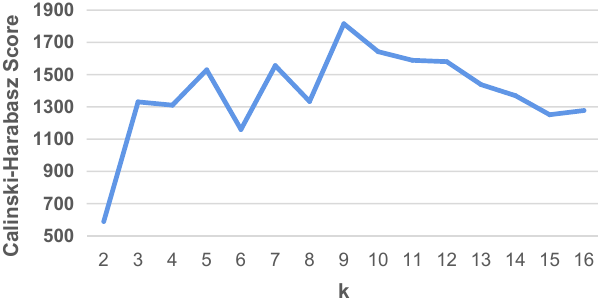}
\caption{Calinski-Harabasz Scores of Different k}
\label{fig_cluster}
\end{figure}

\paragraph{Skill Clustering} In the skill discovery phase, we use the k-means clustering algorithm to cluster subgoals into different groups based on their semantic similarity.
On the choice of $k$, we compute the Calinski-Harabasz (CH) score, which is a commonly used metric for clustering results, from $k=2$ to $16$ as shown in Figure \ref{fig_ch}, and finally chose $k=9$ with the highest CH score.
We also show the clustering results of semantic embedding through the TSNE technique in Figure \ref{fig_cluster}.
However, unsupervised clustering may cause some uncertainty in the skill discovery process, we leave further investigating and minimizing the effect of clustering results on our framework as future works.

\section{6 Conclusion}i24
In this paper, we propose a framework called Self-Driven Grounding (SDG) to address the challenge of automatically grounding LLM onto specific environments. 
In order to alleviate the problem of grounding experience acquisition, we make the LLM not only decompose tasks but also generate intrinsic rewards to help RL agents efficiently verify the decomposition results.
We also propose a language-aligned general skill learning methodology by forcing each skill to achieve a group of goals with similar semantic descriptions to enhance their generality.
Compared with imitation learning methods that cost millions of demonstrations, SDG can achieve comparable performance in the hardest tasks in BabyAI. 
The ablation study also shows the flexibility and feasibility of learned skills in the interactions between the high-level planner and the environment.

However, limited by the simple way of perceiving the environment status, SDG can only deal with tasks with textual descriptions.
This problem can be addressed by introducing multi-modal LLMs, which will extend the applications of SDG.
Besides, SDG only contains a single cycle of hypothesis, verification, and induction.
It is an interesting and promising direction to design a mechanism of multiple cycles in SDG, allowing SDG to learn more powerful and diverse hierarchical skills to accomplish more flexible tasks.


\bibliography{aaai24}

\begin{thebibliography}{25}
\providecommand{\natexlab}[1]{#1}

\bibitem[{hig(2020)}]{higher}
 2020.
\newblock \emph{2020 {IEEE} Symposium Series on Computational Intelligence,
  {SSCI} 2020, Canberra, Australia, December 1-4, 2020}. {IEEE}.
\newblock ISBN 978-1-7281-2547-3.

\bibitem[{Artzi and Zettlemoyer(2013)}]{iw0}
Artzi, Y.; and Zettlemoyer, L. 2013.
\newblock Weakly Supervised Learning of Semantic Parsers for Mapping
  Instructions to Actions.
\newblock \emph{Trans. Assoc. Comput. Linguistics}, 1: 49--62.

\bibitem[{Bahdanau et~al.(2019)Bahdanau, Hill, Leike, Hughes, Hosseini, Kohli,
  and Grefenstette}]{rwirl1}
Bahdanau, D.; Hill, F.; Leike, J.; Hughes, E.; Hosseini, S.~A.; Kohli, P.; and
  Grefenstette, E. 2019.
\newblock Learning to Understand Goal Specifications by Modelling Reward.
\newblock In \emph{7th International Conference on Learning Representations,
  {ICLR} 2019, New Orleans, LA, USA, May 6-9, 2019}.

\bibitem[{Carta et~al.(2023)Carta, Romac, Wolf, Lamprier, Sigaud, and
  Oudeyer}]{llm_finetune}
Carta, T.; Romac, C.; Wolf, T.; Lamprier, S.; Sigaud, O.; and Oudeyer, P. 2023.
\newblock Grounding Large Language Models in Interactive Environments with
  Online Reinforcement Learning.
\newblock \emph{CoRR}, abs/2302.02662.

\bibitem[{Chaplot et~al.(2018)Chaplot, Sathyendra, Pasumarthi, Rajagopal, and
  Salakhutdinov}]{rwil2}
Chaplot, D.~S.; Sathyendra, K.~M.; Pasumarthi, R.~K.; Rajagopal, D.; and
  Salakhutdinov, R. 2018.
\newblock Gated-Attention Architectures for Task-Oriented Language Grounding.
\newblock In \emph{Proceedings of the Thirty-Second {AAAI} Conference on
  Artificial Intelligence, (AAAI-18), the 30th innovative Applications of
  Artificial Intelligence (IAAI-18), and the 8th {AAAI} Symposium on
  Educational Advances in Artificial Intelligence (EAAI-18), New Orleans,
  Louisiana, USA, February 2-7, 2018}, 2819--2826.

\bibitem[{Chen, Gupta, and Marino(2021)}]{rwhr2}
Chen, V.; Gupta, A.; and Marino, K. 2021.
\newblock Ask Your Humans: Using Human Instructions to Improve Generalization
  in Reinforcement Learning.
\newblock In \emph{9th International Conference on Learning Representations,
  {ICLR} 2021, Virtual Event, Austria, May 3-7, 2021}.

\bibitem[{Chevalier{-}Boisvert et~al.(2019)Chevalier{-}Boisvert, Bahdanau,
  Lahlou, Willems, Saharia, Nguyen, and Bengio}]{babyai}
Chevalier{-}Boisvert, M.; Bahdanau, D.; Lahlou, S.; Willems, L.; Saharia, C.;
  Nguyen, T.~H.; and Bengio, Y. 2019.
\newblock BabyAI: {A} Platform to Study the Sample Efficiency of Grounded
  Language Learning.
\newblock In \emph{7th International Conference on Learning Representations,
  {ICLR} 2019, New Orleans, LA, USA, May 6-9, 2019}.

\bibitem[{Driess et~al.(2023)Driess, Xia, Sajjadi, Lynch, Chowdhery, Ichter,
  Wahid, Tompson, Vuong, Yu, Huang, Chebotar, Sermanet, Duckworth, Levine,
  Vanhoucke, Hausman, Toussaint, Greff, Zeng, Mordatch, and Florence}]{ground1}
Driess, D.; Xia, F.; Sajjadi, M. S.~M.; Lynch, C.; Chowdhery, A.; Ichter, B.;
  Wahid, A.; Tompson, J.; Vuong, Q.~H.; Yu, T.; Huang, W.; Chebotar, Y.;
  Sermanet, P.; Duckworth, D.; Levine, S.; Vanhoucke, V.; Hausman, K.;
  Toussaint, M.; Greff, K.; Zeng, A.; Mordatch, I.; and Florence, P.~R. 2023.
\newblock PaLM-E: An Embodied Multimodal Language Model.
\newblock \emph{ArXiv}, abs/2303.03378.

\bibitem[{Garg et~al.(2022)Garg, Vaidyanath, Kim, Song, and Ermon}]{LISA}
Garg, D.; Vaidyanath, S.; Kim, K.; Song, J.; and Ermon, S. 2022.
\newblock LISA: Learning Interpretable Skill Abstractions from Language.
\newblock \emph{ArXiv}, abs/2203.00054.

\bibitem[{Huang et~al.(2022)Huang, Xia, Xiao, Chan, Liang, Florence, Zeng,
  Tompson, Mordatch, Chebotar, Sermanet, Jackson, Brown, Luu, Levine, Hausman,
  and Ichter}]{innermono}
Huang, W.; Xia, F.; Xiao, T.; Chan, H.; Liang, J.; Florence, P.; Zeng, A.;
  Tompson, J.; Mordatch, I.; Chebotar, Y.; Sermanet, P.; Jackson, T.; Brown,
  N.; Luu, L.; Levine, S.; Hausman, K.; and Ichter, B. 2022.
\newblock Inner Monologue: Embodied Reasoning through Planning with Language
  Models.
\newblock In \emph{Conference on Robot Learning, CoRL 2022, 14-18 December
  2022, Auckland, New Zealand}, 1769--1782.

\bibitem[{Ichter et~al.(2022)Ichter, Brohan, Chebotar, Finn, Hausman, Herzog,
  Ho, Ibarz, Irpan, Jang, Julian, Kalashnikov, Levine, Lu, Parada, Rao,
  Sermanet, Toshev, Vanhoucke, Xia, Xiao, Xu, Yan, Brown, Ahn, Cortes, Sievers,
  Tan, Xu, Reyes, Rettinghouse, Quiambao, Pastor, Luu, Lee, Kuang, Jesmonth,
  Joshi, Jeffrey, Ruano, Hsu, Gopalakrishnan, David, Zeng, and Fu}]{saycan}
Ichter, B.; Brohan, A.; Chebotar, Y.; Finn, C.; Hausman, K.; Herzog, A.; Ho,
  D.; Ibarz, J.; Irpan, A.; Jang, E.; Julian, R.; Kalashnikov, D.; Levine, S.;
  Lu, Y.; Parada, C.; Rao, K.; Sermanet, P.; Toshev, A.; Vanhoucke, V.; Xia,
  F.; Xiao, T.; Xu, P.; Yan, M.; Brown, N.; Ahn, M.; Cortes, O.; Sievers, N.;
  Tan, C.; Xu, S.; Reyes, D.; Rettinghouse, J.; Quiambao, J.; Pastor, P.; Luu,
  L.; Lee, K.; Kuang, Y.; Jesmonth, S.; Joshi, N.~J.; Jeffrey, K.; Ruano,
  R.~J.; Hsu, J.; Gopalakrishnan, K.; David, B.; Zeng, A.; and Fu, C.~K. 2022.
\newblock Do As {I} Can, Not As {I} Say: Grounding Language in Robotic
  Affordances.
\newblock In \emph{Conference on Robot Learning, CoRL 2022, 14-18 December
  2022, Auckland, New Zealand}, 287--318.

\bibitem[{Kojima et~al.(2022)Kojima, Gu, Reid, Matsuo, and Iwasawa}]{LLM4}
Kojima, T.; Gu, S.~S.; Reid, M.; Matsuo, Y.; and Iwasawa, Y. 2022.
\newblock Large Language Models are Zero-Shot Reasoners.
\newblock \emph{ArXiv}, abs/2205.11916.

\bibitem[{Li et~al.(2022)Li, Puig, Paxton, Du, Wang, Fan, Chen, Huang,
  Aky{\"{u}}rek, Anandkumar, Andreas, Mordatch, Torralba, and Zhu}]{llm_ft1}
Li, S.; Puig, X.; Paxton, C.; Du, Y.; Wang, C.; Fan, L.; Chen, T.; Huang, D.;
  Aky{\"{u}}rek, E.; Anandkumar, A.; Andreas, J.; Mordatch, I.; Torralba, A.;
  and Zhu, Y. 2022.
\newblock Pre-Trained Language Models for Interactive Decision-Making.
\newblock In \emph{NeurIPS}.

\bibitem[{Liang et~al.(2022)Liang, Huang, Xia, Xu, Hausman, Ichter, Florence,
  and Zeng}]{codepol}
Liang, J.; Huang, W.; Xia, F.; Xu, P.; Hausman, K.; Ichter, B.; Florence, P.;
  and Zeng, A. 2022.
\newblock Code as Policies: Language Model Programs for Embodied Control.
\newblock \emph{CoRR}, abs/2209.07753.

\bibitem[{Lynch and Sermanet(2021)}]{rwil1}
Lynch, C.; and Sermanet, P. 2021.
\newblock Language Conditioned Imitation Learning Over Unstructured Data.
\newblock In \emph{Robotics: Science and Systems XVII, Virtual Event, July
  12-16, 2021}.

\bibitem[{Misra et~al.(2016)Misra, Sung, Lee, and Saxena}]{iw1}
Misra, D.~K.; Sung, J.; Lee, K.; and Saxena, A. 2016.
\newblock Tell me Dave: Context-sensitive grounding of natural language to
  manipulation instructions.
\newblock \emph{Int. J. Robotics Res.}, 35(1-3): 281--300.

\bibitem[{OpenAI(2023)}]{LLM3}
OpenAI. 2023.
\newblock GPT-4 Technical Report.
\newblock \emph{ArXiv}, abs/2303.08774.

\bibitem[{Raman et~al.(2022)Raman, Cohen, Rosen, Idrees, Paulius, and
  Tellex}]{reprompt}
Raman, S.~S.; Cohen, V.; Rosen, E.; Idrees, I.; Paulius, D.; and Tellex, S.
  2022.
\newblock Planning with Large Language Models via Corrective Re-prompting.
\newblock \emph{CoRR}, abs/2211.09935.

\bibitem[{Ranzato et~al.(2021)Ranzato, Beygelzimer, Dauphin, Liang, and
  Vaughan}]{ella}
Ranzato, M.; Beygelzimer, A.; Dauphin, Y.~N.; Liang, P.; and Vaughan, J.~W.,
  eds. 2021.
\newblock \emph{Advances in Neural Information Processing Systems 34: Annual
  Conference on Neural Information Processing Systems 2021, NeurIPS 2021,
  December 6-14, 2021, virtual}.

\bibitem[{R{\"{o}}der, Eppe, and Wermter(2022)}]{rwhr1}
R{\"{o}}der, F.; Eppe, M.; and Wermter, S. 2022.
\newblock Grounding Hindsight Instructions in Multi-Goal Reinforcement Learning
  for Robotics.
\newblock In \emph{{IEEE} International Conference on Development and Learning,
  {ICDL} 2022, London, United Kingdom, September 12-15, 2022}, 170--177.

\bibitem[{Wang et~al.(2023)Wang, Xie, Jiang, Mandlekar, Xiao, Zhu, Fan, and
  Anandkumar}]{voyager}
Wang, G.; Xie, Y.; Jiang, Y.; Mandlekar, A.; Xiao, C.; Zhu, Y.; Fan, L.; and
  Anandkumar, A. 2023.
\newblock Voyager: An Open-Ended Embodied Agent with Large Language Models.
\newblock \emph{CoRR}, abs/2305.16291.

\bibitem[{Wang et~al.(2022)Wang, Jansen, C{\^{o}}t{\'{e}}, and
  Ammanabrolu}]{llm_ft2}
Wang, R.; Jansen, P.~A.; C{\^{o}}t{\'{e}}, M.; and Ammanabrolu, P. 2022.
\newblock ScienceWorld: Is your Agent Smarter than a 5th Grader?
\newblock In \emph{Proceedings of the 2022 Conference on Empirical Methods in
  Natural Language Processing, {EMNLP} 2022, Abu Dhabi, United Arab Emirates,
  December 7-11, 2022}, 11279--11298.

\bibitem[{Wei et~al.(2021)Wei, Bosma, Zhao, Guu, Yu, Lester, Du, Dai, and
  Le}]{LLM2}
Wei, J.; Bosma, M.; Zhao, V.; Guu, K.; Yu, A.~W.; Lester, B.; Du, N.; Dai,
  A.~M.; and Le, Q.~V. 2021.
\newblock Finetuned Language Models Are Zero-Shot Learners.
\newblock \emph{ArXiv}, abs/2109.01652.

\bibitem[{Wei et~al.(2022)Wei, Wang, Schuurmans, Bosma, hsin Chi, Xia, Le, and
  Zhou}]{LLM1}
Wei, J.; Wang, X.; Schuurmans, D.; Bosma, M.; hsin Chi, E.~H.; Xia, F.; Le, Q.;
  and Zhou, D. 2022.
\newblock Chain of Thought Prompting Elicits Reasoning in Large Language
  Models.
\newblock \emph{ArXiv}, abs/2201.11903.

\bibitem[{Yuan et~al.(2023)Yuan, Zhang, Wang, Xie, Cai, Dong, and Lu}]{plan4mc}
Yuan, H.; Zhang, C.; Wang, H.; Xie, F.; Cai, P.; Dong, H.; and Lu, Z. 2023.
\newblock Plan4MC: Skill Reinforcement Learning and Planning for Open-World
  Minecraft Tasks.
\newblock \emph{CoRR}, abs/2303.16563.

\end{thebibliography}
\newpage
\section{A. BabyAI Environment}
\begin{figure}[tb]
\centering
\includegraphics[width=0.48\textwidth]{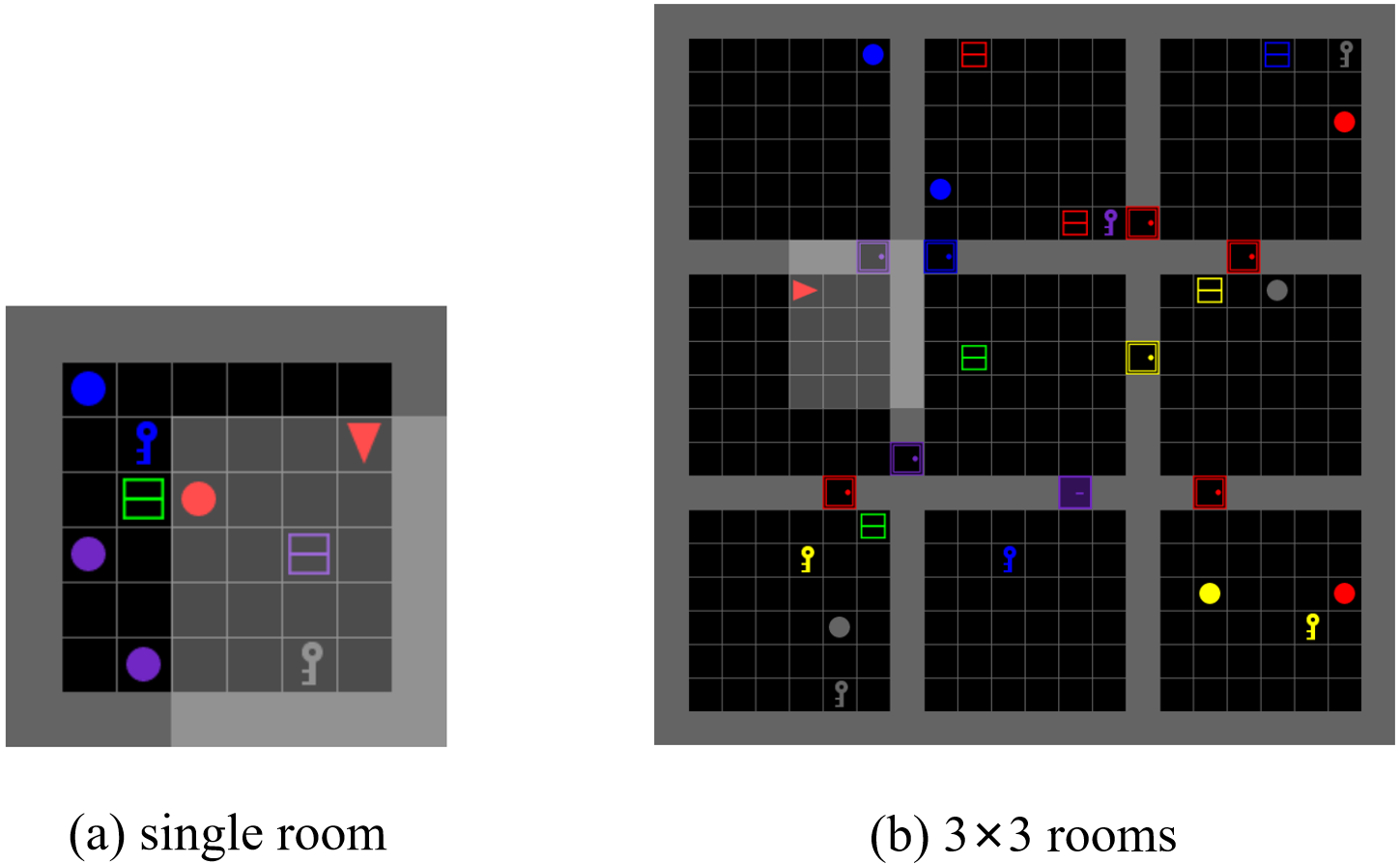}
\caption{Different maps in BabyAI}
\label{fig_maps}
\end{figure}

BabyAI is a grid world environment for instruction following investigations, which consists of 19 levels of tasks with increasing difficulty.
All instructions in BabyAI are generated through a combinatorially rich synthetic grammar.
Considering some levels repeat each other or are too specific to a single scene, like ``GoToRedBall'', We selected 4 representative single-task scenarios and 2 challenging combined-task levels.
There are two kinds of map in BabyAI, single room or $3\times 3$ rooms as shown in Figure \ref{fig_maps}.

The following Table \ref{tb_tasks} shows the details of each level of tasks.
\begin{table*}[]
\centering
\begin{tabular}{@{}llll@{}}
\toprule
Level        & Map               & Scenario      & Mission Space                                                                                                                                 \\ \midrule
GoToLocal    & single room       & single-task   & go to the \{color\} \{type\}                                                                                                                  \\
PickupLoc    & single room       & single-task   & pick up the \{color\} \{type\}                                                                                                                \\
PutNextLocal & single room       & single-task   & put the \{color\} \{type\} next to the \{color\} \{type\}                                                                                     \\
Open         & $3\times 3$ rooms & single-task   & open a \{color\} door                                                                                                                         \\
SynthSeq     & $3\times 3$ rooms & combined-task & \begin{tabular}[c]{@{}l@{}}Combined instructions of "GoTo", "PickUp", "PutNext", "Open" \\ using "and", ", then" and "after you"\end{tabular} \\
BossLevel    & $3\times 3$ rooms & combined-task & Command can be any sentence drawn from the Baby Language grammar                                                                              \\ \bottomrule
\end{tabular}
\caption{Details of 6 selected levels. ``color'' can be ``re'', ``green'', ``blue'', ``purple'', ``yellow'' or ``grey''. ``type'' can be ``ball'', ``box'' or ``key''.}
\label{tb_tasks}
\end{table*}

\section{B. Hypothesis Details}
As described in the paper, in the hypothesis phase, HYVIN makes LLM not only decompose the instruction into subgoals but also corresponding check functions.
The prompt consists of the role definition, task definitions, perception function descriptions, and the output format constraint.

\begin{shaded}
\centerline{Role Definition}
\noindent There is a robot in the maze, which consists of multiple rooms connected by doors.
Some objects are in the room, like boxes, balls, doors, and keys.
You are a kindful assistant that helps to guide the robot to complete some objects manipulating tasks.
Please reply ``Yes'' if you understand your role, and then I will give your more details information.
\end{shaded}

\begin{shaded}
\centerline{Task Definitions}
\noindent Tasks like ``go to X'', the robot should first discover X, and then go next to X.\\
\noindent Tasks like ``pick X'', the robot should first ``go to X'', then pick X.\\
\noindent Tasks like ``put X next to Y'', the robot should first ``pick X'', then discover Y, and finally put X next to Y.\\
\noindent Tasks like ``open the X'',
the robot should consider door X may be in the current room or another room.
The robot should first try to discover, go to, and open door X in the current room.
Otherwise, the robot should go to a new room. Specifically, it should find an unexplored door Y, and then enter the new room door Y leads to.
\end{shaded}

\begin{shaded}
\centerline{Perception Function Descriptions}
You can use the following Python APIs to get the status of the robot.\\
\begin{lstlisting}
(1) get_observation():
# This API returns a dictionary containing all objects in the robot's vision. The key is the names of the objects, the value is the object's coordinates relative to the robot.
# For example:
obs = robot.get_observation()
# output: obs = {"blue key":[-1,2], "yellow box":[0,3]}
# You can use the observation to judge whether the robot has discovered a specific object or a kind of object, like:
if "red ball" in robot.get_observation().keys():
    do_something()
# Or you can judge whether the robot is next to a specific object, like:
obs = robot.get_observation()
if "red ball" in obs.keys()
    if sum(abs(obs["red ball"])) == 1:
        do_something()
# Or you can judge whether one object is next to another object, like:
obs = robot.get_observation()
if "red ball" in obs.keys() and "green key" in obs.keys():
    if sum(abs(obs["red ball"] - obs["green key"])) == 1:
        do_something()
(2) get_carried():
# This API returns the current carried object of the robot.
# For example:
carried_item = robot.get_carried()
# carried_item = "purple ball", means the robot has carried a "purple ball" now. 
# carried_item = None, means the robot has not pick anything now. 
# You can use this API to judge whether the robot has got the target object before putting it in a special place:
if robot.get_carried() == "blue box":
    put_item_to_a_place()
(3) get_visited_rooms():
# This API returns the id list of rooms that have been visited by the robot recently.
# For example:
visited = robot.get_visited_rooms()
# output: visited = [0,7,3] means the robot has visited No. 0, 7, and 3 rooms recently
(4) get_doors_in_observation():
# This API returns a dictionary containing all doors in the robot's observation. Keys are the names of doors, the value is the coordinates relative to the robot, the status (open or closed), and the id of the room this door leads to.
# For example:
doors = robot.get_doors_in_observation()
# output: doors = {"purple door":[2,1,0,3], "green door":[-1,3,1,6]}, the first two items are the coordinates, the third item is the status of the door, the fourth item is the room id.
# You can check every door is open or close:
doors = robot.get_doors_in_observation()
for door in doors.keys():
    status = doors[door][2]
    room_id = doors[door][3]
    if status == 0: # the door is open
        ...
# Or you can discover doors lead to unexplored rooms:
doors = robot.get_doors_in_observation()
for door in doors.keys():
    room_id = doors[door][3]
    if room_id not in robot.get_visited_rooms:
       do_something()
(5) get_current_room()
# The API returns the current room id.
current = robot.get_current_room()
# output: current = 1 means the robot now in No. 1 room.
\end{lstlisting}
\end{shaded}

\begin{shaded}
\centerline{Output Format Constraints}   
After the robot is given the task, your works are:\\
(1) decompose the task instruction into several goals to help the robot complete the task. Each goal should be small and easy to check.\\
(2) write separate Python functions using the above two APIs to check whether the robot achieves each goal. Only check function, and no implementation to help the robot.\\
Your output should follow this format:\\
Task instruction: ``...''\\
Goals:\\
(1) xxx...
\begin{lstlisting}
def checkxxx...(robot):
    if xxx...:
        return True
    else:
        return False
\end{lstlisting}
(2)xxx...
\begin{lstlisting}
def xxx...
\end{lstlisting}
xxx...\\
Please reply ``Yes'' if you understand the above APIs and your works, and then I will give your the object manipulating task."
\end{shaded}

Here are some decomposition examples:
\begin{shaded}
\noindent User: Task instruction: ``put the red key next to the red ball''\\
ChatGPT: Task instruction: ``Put the red key next to the red bal''\\
Goals:\\
(1) Check if the robot has discovered the red key
\begin{lstlisting}
    def check_discovered_red_key(robot):
        if "red key" in robot.get_observation().keys():
            return True
        else:
            return False
\end{lstlisting}
(2) Check if the robot is next to the red key
\begin{lstlisting}
    def check_next_to_red_key(robot):
        obs = robot.get_observation()
        if "red key" in obs.keys():
            if sum(abs(obs["red key"])) == 1:
                return True
        return False
\end{lstlisting}
(3) Check if the robot has picked the red key
\begin{lstlisting}
    def check_picked_red_key(robot):
        if robot.get_carried() == "red key":
            return True
        else:
            return False
\end{lstlisting}

\end{shaded}
\begin{shaded}
\noindent(4) Check if the robot has discovered the red ball
\begin{lstlisting}
    def check_discovered_red_ball(robot):
        if "red ball" in robot.get_observation().keys():
            return True
        else:
            return False    
\end{lstlisting}
(5) Check if the robot is next to the red ball
\begin{lstlisting}
    def check_next_to_red_ball(robot):
        obs = robot.get_observation()
        if "red ball" in obs.keys():
            if sum(abs(obs["red ball"])) == 1:
                return True
        return False
\end{lstlisting}
(6) Check if the robot has put the red key next to the red ball
\begin{lstlisting}
    def check_put_next_to_red_ball(robot):
        obs = robot.get_observation()
        if "red key" in obs.keys() and "red ball" in obs.keys():
            if sum(abs(obs["red key"] - obs["red ball"])) == 1:
                return True
        return False
\end{lstlisting}
Note: These functions are only for checking the completion of each goal. The implementation of the robot's actions to achieve the goals is not included.\\
Please let me know if you need any further assistance.
\end{shaded}
\begin{shaded}
\noindent User: Task instruction: ``open the green door''\\
ChatGPT: Goals:\\
Goal 1: Locate the current room's doors and objects
\begin{lstlisting}
def goal_1_locate_objects(robot):
    observation = robot.get_observation()
    return True if observation else False
\end{lstlisting}
Goal 2: Check for the green door in the current room
\begin{lstlisting}
def goal_2_find_green_door(robot):
    observation = robot.get_observation()
    return True if "green door" in observation.keys() else False
\end{lstlisting}
Goal 3: Approach and open the green door
\begin{lstlisting}
def goal_3_approach_and_open_door(robot):
    doors_in_observation = robot.get_doors_in_observation()
    return True if "green door" in doors_in_observation.keys() and doors_in_observation["green door"][2] == 1 else False    
\end{lstlisting}
Goal 4: Find an unexplored door in the current room
\begin{lstlisting}
def goal_4_find_unexplored_door(robot):
    doors_in_observation = robot.get_doors_in_observation()
    for door in doors_in_observation.keys():
        room_id = doors_in_observation[door][3]
        if room_id not in robot.get_visited_rooms():
            return True
    return False    
\end{lstlisting}
Goal 5: Enter the unexplored room
\begin{lstlisting}
def goal_5_enter_unexplored_room(robot):
    current_room = robot.get_current_room()
    visited_rooms = robot.get_visited_rooms()
    return True if current_room not in visited_rooms else False   
\end{lstlisting}
Goal 6: Repeat from Goal 1 in the new room\\
This goal is already covered by the looping nature of the task instructions. We don't need a separate function for this goal.\\

Note: These functions only check whether the robot has achieved each specific goal based on the given API information. They don't include the actual task implementation or navigation instructions for the robot. The navigation part would depend on the higher-level program that calls these functions and guides the robot through the maze.
\end{shaded}

\section{C. Deduction Details}
\subsection{Generation Prompt}
Similar to the decomposition prompt, the generation prompt also has role definition, perception function descriptions (same as the decomposition prompt), and output format constraints.
Besides, considering the difficulties when faced with unseen and long-term tasks, we add skill descriptions and an example to enable the in-context learning mechanism.
\begin{shaded}
    \centerline{Role Definition}
\noindent There is a robot in the maze, which consists of multiple rooms connected by doors. Some objects are in the room, like boxes, balls, doors, and keys. You are a kindful assistant that helps to generate the Python program to control the robot to complete some objects manipulating tasks
\end{shaded}
\begin{shaded}
    \centerline{Skill Descriptions}
\noindent You can use the following Python APIs to control the robot or get the state of the robot:
\begin{lstlisting}
(1) discover_object(object_name):
# This API makes the robot discover a specific object (including keys, boxes, or balls). If there is no such object in the current room, it may fail and the robot needs to explore another room.
# For example:
robot.discover_object("blue ball")
if "blue ball" in robot.get_observation().keys():
    do_somthing_1() # It means the robot has discovered the "blue ball"
else:
    do_somthing_2()
(2) go_next_to(object_name):
# When the robot has discovered the target object, this API makes the robot go next to the object.
# For example:
if "blue ball" in robot.get_observation().keys():
    robot.go_next_to("blue ball")
(3) pick(object_name):
# When the robot has been next to the target object, this API makes the robot pick the object.
# For example:
if sum(abs(robot.get_observation["green box"])) == 1: # the green box is next to the robot
    robot.pick("green box")
(4) put_next_to(object_name1, object_name2):
# When the robot has been next to the target object 2 and has picked object 1, this API makes the robot put object 1 next to object 2.
# For example:
if sum(abs(robot.get_observation["green box"])) == 1: # the green box is next to the robot
    if "purple key" in robot.get_picked(): # The robot has picked the purple key
    robot.put_next_to("purple key", "green box")
(5) discover_door(door_name):
# This API makes the robot discover a specific door. If there is no such door in the current room, it may fail.
# For example:
robot.discover_door("blue door")
if "blue door" in robot.get_observation().keys():
    do_somthing_1() # It means the robot has discovered the "blue door"
(6) go_to_door(door_name):
# When the robot has discovered the target door, this API makes the robot go to the door.
# For example:
if "red door" in robot.get_observation().keys():
    robot.go_to_door("red door")
(7) open_door(door_name):
# This API makes the robot open the door.
# For example:
doors = robot.get_doors_in_observation()
if "red door" in door.keys(): # doors in the robot's observation
    if sum(doors["red door"][:2]) == 1: the door is in front of the robot
        robot.open_door("red door")
(8) find_unexplored_door():
# When the robot cannot discover a specific object in the current room, you can use this API to guide the robot to find a door that leads to an unexplored room.
(9) enter_unexplored_room():
# When the robot has found an unexplored door, this API guides the robot to open the door and enter an unexplored room.
# For example:
robot.discover_object("grey box")
while not  "grey box" in robot.get_observation().keys():
    robot.find_unexplored_door()
    robot.enter_unexplored_room()
    robot.discover_object("grey box")
 # When the while loop ends, it means the robot has discovered the "grey box"
\end{lstlisting}
\end{shaded}
\begin{shaded}
    \centerline{Example}
\noindent Here is an example of a good program:\\
Instruction: put the blue ball on your left next to a green ball after you put a blue box next to the yellow door and open the grey door\\
\begin{lstlisting}
# Program begin:
# task 1: put a blue box next to the yellow door
def put_blue_box_next_to_yellow_door(robot): 
    # 1. pick the blue box
    # discover the blue box
    robot.discover_object("blue box")
    while "blue box" not in robot.get_observation().keys():
        # if cannot discover the blue box, explore the next room
        res = robot.find_unexplored_door()
        robot.enter_unexplored_room()
        robot.discover_object("blue box")
    # flush the visited room trajectory flush when discover blue box
    robot.flush()
    # go to the blue box
    robot.go_next_to("blue box")
    # pick the blue box
    robot.pick("blue box")

    # 2. put blue box next to yellow box
    # discover the yellow door
    robot.discover_door("yellow door")
    while "yellow door" not in robot.get_observation().keys():
        # if cannot discover the yellow door, explore the next room
        res = robot.find_unexplored_door()
        robot.enter_unexplored_room()
        robot.discover_door("yellow door")
    # flush the visited room trajectory flush when discover yellow box
    robot.flush()
    # go to yellow door
    robot.go_next_to_door("yellow door")
    # put
    robot.put_next_to("blue box", "yellow door")
    return

# task 2: open the grey door
def open_grey_door(robot):
    # 1. open the grey door
    # discover the grey door
    robot.discover_door("grey door")
    while "grey door" not in robot.get_observation().keys():
        # if cannot discover the grey door, explore the next room
        robot.find_unexplored_door()
        robot.enter_unexplored_room()
        robot.discover_door("grey door")
    # flush the visited room trajectory flush when discover grey door
    robot.flush()
    # go to grey door
    robot.go_next_to_door("grey door")
    # open grey door
    robot.open_door("grey door")
    return

# task 3: put the blue ball next to a green ball
def put_blue_ball_next_to_green_ball(robot):
    # 1. pick the blue ball
    # discover the blue ball
    robot.discover_object("blue ball")
    while "blue ball" not in robot.get_observation().keys():
        # if cannot discover the blue ball, explore the next room
        robot.find_unexplored_door()
        robot.enter_unexplored_room()
        robot.discover_object("blue ball")
    # flush the visited room trajectory flush when discover blue ball
    robot.flush()
    # go to the blue ball
    robot.go_next_to("blue ball")
    # pick the blue ball
    robot.pick("blue ball")

    # 2. put blue ball next to green ball
    # discover the green ball
    robot.discover_door("green ball")
    while "green ball" not in robot.get_observation().keys():
        # if cannot discover the green ball, explore the next room
        res = robot.find_unexplored_door()
        robot.enter_unexplored_room()
        robot.discover_door("green ball")
    # flush the visited room trajectory flush when discover green ball
    robot.flush()
    # go to green ball
    robot.go_next_to("green ball")
    # put
    robot.put_next_to("blue ball", "green ball")
    return


# main:
put_blue_box_next_to_yellow_door(robot)
open_grey_door(robot)
put_blue_ball_next_to_green_ball(robot)
# Program end
\end{lstlisting}
If you understand the example and your work, please wait for the task instruction.
\end{shaded}
\begin{shaded}
    \centerline{Output Format Constraints}
\noindent As the example shows, the whole task may consists mutiple small task or only one small task.\\
Your output should follow this format:\\
Instruction: xxx\\
\begin{lstlisting}    
# Program begin:
# task 1: xxxx
def task1_xxx...(robot):
    xxx...
# task 2: xxx...
xxx...
# taskxxx...
...
# main
task1_xxx...(robot)
task2_xxx...(robot)
...
# Program end
\end{lstlisting}
\end{shaded}
Here are some generation examples:
\begin{shaded}
\noindent Instruction: put the grey ball next to the red key
\begin{lstlisting}
# Program begin:
# task 1: put grey ball next to the red key
def put_grey_ball_next_to_red_key(robot):
    # 1. pick the grey ball
    # discover the grey ball
    robot.discover_object("grey ball")
    # go to the grey ball
    robot.go_next_to("grey ball")
    # pick the grey ball
    robot.pick("grey ball")

    # 2. put grey ball next to red key
    # discover the red key
    robot.discover_object("red key")
    # go to red key
    robot.go_next_to("red key")
    # put
    robot.put_next_to("grey ball", "red key")
    return

# main
put_grey_ball_next_to_red_key(robot)
# Program end
\end{lstlisting}
\end{shaded}

\begin{shaded}
\noindent Instruction: Put a ball next to the box on your left and open a door after you open a green door and put the yellow box on your left next to a red door. 
\begin{lstlisting}
# Program begin:
# Task 1: Open the green door
def open_green_door(robot):
    # Discover the green door
    robot.discover_door("green door")
    
    while "green door" not in robot.get_observation().keys():
        # If the green door cannot be discovered, explore the next room
        robot.find_unexplored_door()
        robot.enter_unexplored_room()
        robot.discover_door("green door")
    
    # Flush the visited room trajectory when discovering the green door
    robot.flush()
    
    # Go to the green door
    robot.go_to_door("green door")
    
    # Open the green door
    robot.open_door("green door")

# Task 2: Put the yellow box on your left next to a red door
def put_yellow_box_next_to_red_door(robot):
    # Discover the yellow box
    robot.discover_object("yellow box")
    
    while "yellow box" not in robot.get_observation().keys():
        # If the yellow box cannot be discovered, explore the next room
        robot.find_unexplored_door()
        robot.enter_unexplored_room()
        robot.discover_object("yellow box")
    
    # Flush the visited room trajectory when discovering the yellow box
    robot.flush()
    
    # Go to the yellow box
    robot.go_next_to("yellow box")
    
    # Pick the yellow box
    robot.pick("yellow box")
    
    # Discover the red door
    robot.discover_door("red door")
    
    while "red door" not in robot.get_observation().keys():
        # If the red door cannot be discovered, explore the next room
        robot.find_unexplored_door()
        robot.enter_unexplored_room()
        robot.discover_door("red door")
    
    # Flush the visited room trajectory when discovering the red door
    robot.flush()
    
    # Go to the red door
    robot.go_to_door("red door")
    
    # Put the yellow box next to the red door
    robot.put_next_to("yellow box", "red door")

# Main program
open_green_door(robot)
put_yellow_box_next_to_red_door(robot)
\end{lstlisting}
\end{shaded}
\subsection{Debugging Prompt}
The debugging prompt is very similar to the generation prompt, except for role definition and output format constraints.
We give the same example as the generation prompt to debugging prompt.
\begin{shaded}
    \centerline{Role Definition}
\noindent There is a robot in the maze, which consists of multiple rooms connected by doors. Some objects are in the room, like boxes, balls, doors, and keys. You are a kindful assistant that helps to modify the given candidate python program of the robot to complete some object manipulating tasks.
\end{shaded}
\begin{shaded}
    \centerline{Output format constraints}
\noindent You would be given the objects manipulating tasks instruction, the candidate program, and some error message, you need to modify the program to guide the robot to complete the task.\\
You need to modify the program to guide the robot to complete the task.\\
You should learn from the above good example and APIs, such as:\\
(1) Before manipulating some object (like open, pick, put, and open door), the robot must discover it first (using some discover APIs) and then go to the target object .\\
(2) For a task like "put A next to the B", before discovering and going to B to put A, the robot must first discover and pick A.\\
You will get the following information:\\
Instruction: xxx\\
\begin{lstlisting}
# Candidate program:
...
# Error messages:
...
\end{lstlisting}
And your output should follow this format:
\begin{lstlisting}
# Modified program begin:
# task 1: xxxx
def task1_xxx...(robot):
    xxx...
# task 2: xxx...
xxx...
# taskxxx...
...
# main
task1_xxx...(robot)
task2_xxx...(robot)
...
# Modified program end
\end{lstlisting}
\end{shaded}
Here is a debugging example:
\begin{shaded}
\noindent User:\\
Instruction: put the red key next to the yellow key\\
\begin{lstlisting}
# Candidate program:
# task 1: put the red key next to the yellow key
def put_red_key_next_to_yellow_key(robot):
    # 1. discover the red key and yellow key
    robot.discover_object("red key")
    robot.discover_object("yellow key")
    
    # 2. go to the red key
    robot.go_next_to("red key")
    
    # 3. pick the red key
    robot.pick("red key")
    
    # 4. go to the yellow key
    robot.go_next_to("yellow key")
    
    # 5. put the red key next to the yellow key
    robot.put_next_to("red key", "yellow key")
    return

# main
put_red_key_next_to_yellow_key(robot)
# Error messages: Pick error!
\end{lstlisting}
ChatGPT:
\begin{lstlisting}
# Modified program begin:
# task 1: put the red key next to the yellow key
def put_red_key_next_to_yellow_key(robot):
    # 1. discover the red key
    robot.discover_object("red key")
    
    # 2. go to the red key
    if "red key" in robot.get_observation().keys():
        robot.go_next_to("red key")
    else:
        raise ValueError("The robot didn't discover the red key before go to it")
    
    # 3. pick the red key
    if sum(abs(robot.get_observation()["red key"])) == 1:
        robot.pick("red key")
    else:
        raise ValueError("The robot is not next to the red key")
    
    # 4. discover the yellow key
    robot.discover_object("yellow key")
    
    # 5. go to the yellow key
    if "yellow key" in robot.get_observation().keys():
        robot.go_next_to("yellow key")
    else:
        raise ValueError("The robot didn't discover the yellow key before go to it")
    
    # 6. put the red key next to the yellow key
    if sum(abs(robot.get_observation()["yellow key"])) == 1:
        robot.put_next_to("red key", "yellow key")
    else:
        raise ValueError("The robot is not next to the yellow key")
    return
    
# main
put_red_key_next_to_yellow_key(robot)
# Modified program end
\end{lstlisting}
\end{shaded}
\section{D. HYVIN-action Prompt}
To demonstrate the effectiveness of self-driven learned skills, we test the variant of our framework, HYVIN-action, which maintains the same high-level planner but employs primitive actions instead of the acquired skills.
The following are the action descriptions and the example in the generation prompt of HYVIN-action.
\begin{shaded}
    \centerline{Action Descriptions}
\noindent You can produce python code to control the robot using the following APIs:
\begin{lstlisting}
(1) turn_left(), turn_right():
# These two APIs can be called without any prerequisites, and it turn the facing direction of the robot and returns observation of the new facing direction.
# For example:
obs_old = robot.get_observation()
obs_new = robot.turn_left()
# obs1 = {"yellow door":[0,3], "purple ball":[-2,1], "green key":[1,0]}
# obs2 = {"yellow door":[3,0], "purple ball":[1,2], "blue box":[-2,0]}
# After turn left, some old seen objects' relative coordinates change (like "yellow door" and "purple ball"), some old objects cannot be seen anymore (like "green key"), and there are some new seen objects.
(2) move_forward():
# This API makes the robot move forward 1 step. It can be called successfully and return true, if there is no object one step forward the robot.
# For a successful example:
obs = robot.get_observation()
state = robot.move_forward()
(3) pick():
# This API makes the robot pick the object at 1 step forward position. It can be called successful if there is an object 1 step forward from the robot and the robot carried nothing now.
# For a successful example:
obs = robot.get_observation() 
assert [0,1] in obs.values and robot.get_carried() is None
state = robot.pick()
new_obs = robot.get_observation()
assert [0,1] not in new_obs.values and robot.get_carried() is not None
(4) put():
# This API makes the robot put the object it carried now in the 1 step forward position. It can be called successful if there is no object 1 step forward the robot and the robot carried one object now.
# For successful example:
obs = robot.get_observation()
assert [0,1] not in obs.values and robot.get_carried() is not None
state = robot.put()
new_obs = robot.get_observation()
assert [0,1] not in new_obs.values and robot.get_carried() is None
(5) toggle()
# This API makes the robot toggle the door at 1 step forward position.
doors_old = robot.get_doors_in_observation()
# doors_old = {"green door":[0,1,1,6]}, the green door is closed
robot.toggle()
doors_new = robot.get_doors_in_observation()
# doors_new = {"green door":[0,1,0,6]}, the green door is open
\end{lstlisting}
\end{shaded}
\begin{shaded}
    \centerline{Example}
Instruction: go to the blue box
\begin{lstlisting}
# Step1:
# Turn facing direction until discovering red ball
obs = robot.get_observation()
for _ in range(4):
    if "blue box" in obs.keys():
        break
    robot.turn_left()
    obs = robot.get_observation()

# Step2:
# Go to the position next to blue box
x_distance, y_distance = obs["blue box"]
# target position is ahead, move forward
if y_distance > 0:
    for _ in range(abs(y_distance)):
        state = robot.move_forward()
# target position is on the right, turn right and move forward 
if x_distance > 0:
    robot.turn_right()
    for _ in range(abs(x_distance)):
        state = robot.move_forward()
# target position is on the left, turn left and move forward 
elif x_distance < 0:
    _ = robot.turn_left()
    for _ in range(abs(x_distance)):
        state = robot.move_forward()
\end{lstlisting}
\end{shaded}
Because of the limited expressive power of the primitive action set, HYVIN-action can only generate simple code logic, which does not perform well on complex tasks.
The following is an example of complex tasks:
\begin{shaded}
\noindent User:\\
Instruction: open a red door and go to the ball on your left after you put the grey ball next to a door.\\
ChatGPT:\\
Certainly, I can help you break down this task into steps and provide you with the Python code to achieve it. Here's the solution:
\begin{lstlisting}
# Step 1
def discover_RedDoor(robot):
    obs = robot.get_observation()
    while "red door" not in obs.keys():
        obs = robot.turn_left()

# Step 2
def open_RedDoor(robot):
    obs = robot.get_observation()
    if "red door" in obs.keys():
        state = robot.step_forward()

# Step 3
def turn_to_GreyBall(robot):
    _ = robot.turn_left()

# Step 4
def approach_GreyBall(robot):
    obs = robot.get_observation()
    x_distance, y_distance = obs["grey ball"]
    
    if y_distance > 0:  # target position is ahead
        for _ in range(y_distance):
            state = robot.step_forward()
    elif x_distance > 0:  # target position is on the right 
        _ = robot.turn_right()
        for _ in range(x_distance):
            state = robot.step_forward()
    elif x_distance < 0:  # target position is on the left
        _ = robot.turn_left()
        for _ in range(abs(x_distance)):
            state = robot.step_forward()

# Step 5
def pick_GreyBall(robot):
    state = robot.pick()

# Step 6
def turn_around(robot):
    for _ in range(2):
        _ = robot.turn_left()

# Step 7
def put_GreyBall_next_to_Door(robot):
    obs = robot.get_observation()
    if "grey ball" in obs.keys():
        state = robot.step_forward()

# Step 8
def turn_to_Ball(robot):
    _ = robot.turn_left()

# Step 9
def approach_Ball(robot):
    obs = robot.get_observation()
    x_distance, y_distance = obs["ball"]
    
    if y_distance > 0:  # target position is ahead
        for _ in range(y_distance):
            state = robot.step_forward()
    elif x_distance > 0:  # target position is on the right 
        _ = robot.turn_right()
        for _ in range(x_distance):
            state = robot.step_forward()
    elif x_distance < 0:  # target position is on the left
        _ = robot.turn_left()
        for _ in range(abs(x_distance)):
            state = robot.step_forward()

# Implementing the solution
def solve_task(robot):
    discover_RedDoor(robot)
    open_RedDoor(robot)
    turn_to_GreyBall(robot)
    approach_GreyBall(robot)
    pick_GreyBall(robot)
    turn_around(robot)
    put_GreyBall_next_to_Door(robot)
    turn_to_Ball(robot)
    approach_Ball(robot)


# Execute the solution
solve_task(robot)
\end{lstlisting}
\end{shaded}
As the example shows, without skills, the LLM can only imitate the given example to call primitive actions directly, thus cannot ground the plan to successful implementation in the environment.

\begin{table*}[]
\centering
\begin{tabular}{@{}cccc@{}}
\toprule
Parameters      & Explanation                                 & Verification & Skill Learning \\ \midrule
max-frames      & The maximum training steps                  & 3e3          & 1e6            \\
discount        & The discount facor or RL                    & \multicolumn{2}{c}{0.99}      \\
lr              & The Learning rate                           & \multicolumn{2}{c}{0.001}     \\
gae-lambda      & The lambda coefficient in GAE formula       & \multicolumn{2}{c}{0.95}      \\
entropy-coef    & The entropy term coefficient                & 0.05         & 0.01           \\
value-loss-coef & The value loss term coefficient             & \multicolumn{2}{c}{0.5}       \\
max-grad-norm   & The maximum norm of gradient                & \multicolumn{2}{c}{0.5}       \\
clip-eps        & The clipping epsilon for PPO                & \multicolumn{2}{c}{0.2}       \\
horizon         & The max steps per episode                   & 30           & 15             \\
drop            & The drop rate of the dropout layer          & 0            & 0.2            \\
text            & Add a GRU to the model to handle text input & \multicolumn{2}{c}{True}      \\ \bottomrule
\end{tabular}
\caption{The parameters of PPO algorithm}
\label{tb_algo}
\end{table*}
\section{E. Reinforcement Learning Details}
In the verification process of the hypothesis phase and skill learning process in the induction phase, we use the standard PPO algorithm and model architecture proposed in BabyAI paper to train the policy.
It is worth noting that in order to promote the generalization ability of learned skills, we add a Dropout layer after the feature extraction in the model.
Table \ref{tb_algo} shows the parameters of reinforcement learning.
\end{document}